\documentclass[letterpaper, 10 pt, conference]{ieeeconf}  

\IEEEoverridecommandlockouts   
\usepackage{graphicx}
\usepackage{amssymb}
\usepackage{amsmath}
\usepackage{changes}
\usepackage{xcolor}
\usepackage{colortbl}
\usepackage[table]{xcolor}
\usepackage{booktabs}

\usepackage{draftwatermark}
\SetWatermarkText{Accepted at IEEE RO-MAN 2026}
\SetWatermarkScale{0.3}

\newcommand{\EditSS}[1]{{\color{black} #1}}

\usepackage[
backend=biber,
style=numeric,
sorting=none
]{biblatex}

\title{\LARGE \bf
Belief-Aware VLM Model for Human-like Reasoning 
}

\author{Anshul Nayak*, Shahil Shaik* and Yue Wang
\thanks{*Equal Contribution}
\thanks{ Anshul Nayak, Shahil Shaik, and Yue Wang are with the Mechanical Engineering
    Department, Clemson University.}%
}

\begin{document}

\maketitle
\thispagestyle{empty}
\pagestyle{empty}

\begin{abstract}

Traditional neural network models for intent inference rely heavily on observable states and struggle to generalize across diverse tasks and dynamic environments. Recent advances in Vision–Language Models (VLMs) and Vision–Language–Action (VLA) models introduce common-sense reasoning through large-scale multimodal pretraining, enabling zero-shot performance across tasks. However, these models still lack explicit mechanisms to represent and update belief, limiting their ability to reason like humans  or capture the  evolving human intent over long-horizon. To address this, we propose a belief-aware VLM framework that integrates retrieval-based memory and reinforcement learning. Instead of learning an explicit belief model, we approximate belief using a vector-based memory that retrieves relevant multimodal context, which is incorporated into the VLM for reasoning. We further refine decision-making using a reinforcement learning policy over the VLM latent space. We evaluate our approach on publicly available VQA datasets such as HD-EPIC and demonstrate consistent improvements over zero-shot baselines, highlighting the importance of belief-aware reasoning.

\end{abstract}

\section{INTRODUCTION}

Robotic systems operating in human-centric environments must reason under uncertainty, anticipate future actions, and adapt to partially observable, dynamically evolving contexts. Classical frameworks such as Markov Decision Processes (MDPs) and Partially Observable MDPs (POMDPs) explicitly model latent states and belief updates, enabling principled reasoning under uncertainty. These formulations have also inspired cognitive models of human reasoning, where agents maintain and update beliefs about others’ intents to guide decision-making. Such belief-driven reasoning is fundamental in both human–human and human–robot interactions, where actions depend not only on the physical state of the environment but also on inferred intentions, expectations, and uncertainty about other agents.

Despite their theoretical rigor, traditional cognitive and probabilistic models rely on hand-crafted representations and simplifying assumptions that limit scalability in high-dimensional, multimodal settings. In contrast, modern deep learning approaches excel at perception but lack explicit mechanisms for structured reasoning and belief modeling. As a result, existing methods often depend on observable states, struggle with long-horizon reasoning, and generalize poorly in complex multi-agent scenarios.

Vision–Language Models (VLMs) have recently emerged as powerful multimodal reasoning systems that integrate visual perception with semantic understanding. Pretrained on large-scale data, they enable strong zero-shot and few-shot generalization and can infer high-level context, relationships, and implicit goals from raw observations, making them well-suited for modeling human behavior and intent.

However, directly applying VLMs in a zero-shot manner is insufficient for modeling nuanced human decision-making in interactive settings. Human behavior and decision-making is inherently belief-driven; individuals continuously update their beliefs about other agents based on observations and use these beliefs to guide their actions. Prior work in cognitive science and robotics has demonstrated the importance of such belief modeling for tasks including intention inference, theory of mind, and collaborative planning. Yet, existing VLM-based approaches largely overlook this aspect, treating reasoning as a static mapping from observations to actions without explicitly modeling the underlying belief dynamics to achieve human-like reasoning.

To address these limitations, we propose a \textbf{belief-aware Vision–Language Model (VLM)} that integrates belief into multimodal reasoning. We model decision-making as a belief-conditioned process, where actions depend on both current observations and prior context. Instead of learning an explicit parametric belief model, we construct a vector-based memory of past multimodal embeddings and retrieve the top-$K$ most relevant contexts based on similarity to approximate belief. This retrieval-based formulation enables context-aware reasoning without explicitly modeling belief dynamics.

The belief mechanism provides prior contextual grounding, enabling the VLM to form richer representations of the environment and agent intent. However, VLMs alone may lack consistency and goal-directed behavior. To address this, we refine the model using a reinforcement learning (RL) policy that optimizes action selection via task-specific rewards. In this framework, belief enhances reasoning, while RL ensures alignment with task objectives. We evaluate our approach on VQA benchmarks such as EPIC-Kitchens, NExT-QA, and HD-EPIC, demonstrating improved performance over zero-shot baselines and highlighting the effectiveness of combining belief-aware representations with reward-driven reasoning.

Our key contributions are summarized as follows:

\begin{itemize}

\item 
We propose a novel \textbf{Belief-aware VLM} framework that incorporates belief into Vision--Language Models through a vector-based memory, where relevant prior multimodal embeddings are retrieved using similarity search to approximate latent belief without requiring an explicit parametric belief model. This enables the VLM to perform context-aware reasoning and approximate intent inference.

\item We augment VLM's contextual reasoning capabilities with  reinforcement learning based decision-making and reasoning through reward-driven optimization.

\item We evaluate our model on popular Vision Query Answer (VQA) datasets such as HD-EPIC \cite{11094504} and show improvement over zero-shot inference performance using state-of-the-art VLMs and other baselines.

\end{itemize}

\section{Related Work}

\subsection{Intent Modeling}

Understanding human preferences and intentions  is critical for effective task execution, transparency, and trust. Prior work has approached intent inference using probabilistic and learning-based methods, including Hidden Markov Models (HMMs) \cite{PETKOVIC2019182} and Dynamic Bayesian Networks (DBNs) \cite{xu2015optimo, hao2025joint} for modeling temporal dependencies, as well as Partially Observable Markov Decision Processes (POMDPs) \cite{zhao2022coordination, bai2015intention} for belief-based reasoning under uncertainty. Supervised approaches, such as Support Vector Machines (SVMs) \cite{kang2015human} and Neural Networks (NNs) \cite{hu2022augmented}, have also been used to predict intent directly from observed features.

Despite their success, these methods often rely on observable states and struggle to generalize across diverse tasks and environments. They are limited by insufficient world knowledge, weak incorporation of language and contextual cues, and short context horizons that hinder modeling of long-term dependencies and evolving goals.

\subsection{Cognitive Modeling With LLM}
Recent advances in Vision–Language Models (VLMs) and Vision–Language–Action (VLA) models have enabled strong zero-shot reasoning by leveraging large-scale multimodal pretraining. Models such as PaLM-E \cite{driess2023palm} and RT-2 \cite{brohan2023rt2visionlanguageactionmodelstransfer} demonstrate that grounding language in perception allows reasoning over physical environments and actions. However, despite exhibiting elements of common-sense reasoning, these approaches lack explicit mechanisms to model belief, intent, and evolving context, limiting their ability to capture long-horizon dependencies and dynamic interactions.

In contrast, cognitive models explicitly incorporate latent belief and intent. Early work on Theory of Mind (ToM) \cite{pmlr-v80-rabinowitz18a} introduced architectures for inferring hidden mental states, while recent studies \cite{zhu2024language} show that LLMs implicitly encode belief representations in their internal activations. Additional efforts integrating cognitive frameworks with LLMs \cite{wu2024cognitive} and alignment methods such as RLHF \cite{havrilla2024teaching} further improve reasoning consistency, highlighting the importance of structured, belief-aware representations for human-like reasoning.

\subsection{RL for LLM reasoning}

Building on these limitations, adapting large VLMs and LLMs to human–robot interaction requires both efficient model adaptation and improved decision-making. Parameter-efficient fine-tuning (PEFT) methods \cite{ding2023parameter, hu2022lora} address scalability by updating only a small subset of parameters while keeping the pretrained backbone fixed, enabling efficient alignment of multimodal representations with task-specific inputs. However, while PEFT improves representation learning, it does not explicitly optimize reasoning or action selection.

To address this, reinforcement learning (RL) has been used to refine model outputs by optimizing task-specific rewards. Approaches such as reinforcement learning from AI feedback (RLAIF) \cite{lee2023rlaif}, improve reasoning by aligning model outputs with desired behaviors, including correctness, consistency, and task completion. This is particularly important in embodied settings, where reasoning must be grounded in both perception and action, motivating the integration of reward-driven optimization with multimodal reasoning models.

Despite these developments, most LLM and VLM approaches still do not explicitly incorporate belief dynamics, which is central to human cognition. Bridging this gap requires integrating structured belief representations with multimodal reasoning. In this work, we address this limitation by incorporating belief-aware reasoning into a VLM framework and refining decision-making through reinforcement learning, enabling more human-like inference grounded in both perception and interaction.

\begin{figure*}
    \centering  \includegraphics[width=0.85\linewidth]{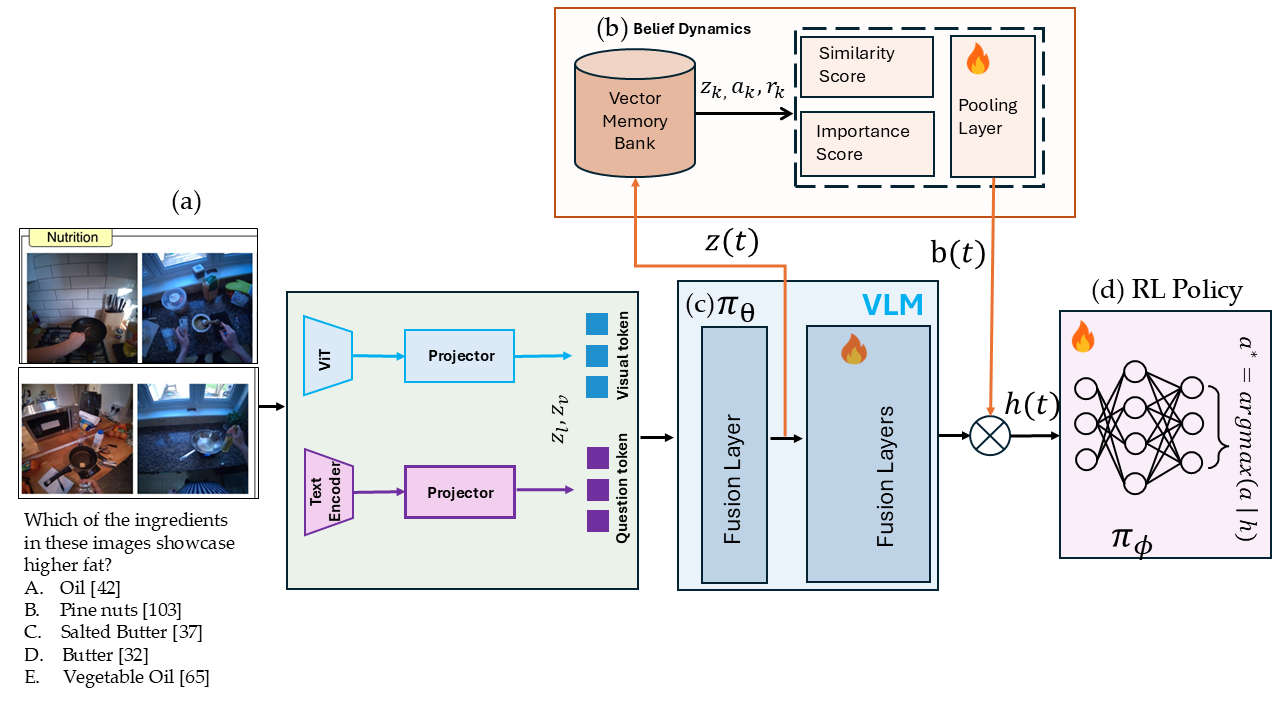}
    \caption{Schema of the proposed Belief-aware VLM. (a) Vision–language inputs encode scene and query context.
(b) A retrieval-based belief module that forms a belief context from past memory.
(c) The VLM $\pi_{\theta}$ fuses multimodal and belief tokens to produce a latent representation.
(d) A policy network refines actions via reinforcement learning.}
    \label{fig:placeholder}
\end{figure*}






\EditSS{\section{Preliminaries}

We consider a multiple-choice video question answering (VQA) setting in which the model must answer a natural-language question about a video by selecting one candidate from a finite answer set. Let $v$ denote a video clip of T timesteps, $\ell$ a task-contextualized natural-language question, and $\mathcal{A}=\{a_1,\dots,a_K\}$ a set of $K$ candidate answers. The goal is to predict the correct answer index $y \in \{1,\dots,K\}$. More generally, this setting can be viewed as decision-making under partial observability, where the current observation may not fully specify the latent factors needed to infer the correct answer. In our setting, these latent factors correspond to unobserved contextual cues and prior situational information that may influence human intent and decision-making.

\subsection{Vision--Language Representation Learning}

Given multimodal inputs, a pretrained VLM maps visual and textual information into a shared latent representation 
\begin{equation}
h = \pi_{\theta}(v, \ell, \mathcal{A})
\end{equation}
where $\theta$ denotes the VLM parameters. The latent representation $h \in \mathbb{R}^{d}$ summarizes the multimodal input and serves as the state used for downstream prediction or policy learning. Depending on the training regime, the encoder $\pi_{\theta}$ may be frozen, adapted with PEFT, or updated end-to-end.

\subsection{Retrieval-Augmented Context Modeling}

To incorporate relevant prior context, we consider an external memory bank (vector database)
\begin{equation}
\mathcal{M} = \{(z_k, a_k, r_k)\}_{k=1}^{N},
\end{equation}
where $z_k$ denotes a stored embedding and $a_k$ denotes the action taken by RL policy and $r_k$ represents the corresponding reward. Given a query embedding $u$, the model retrieves the top-$K$ nearest entries according to a similarity metric:
\begin{equation}
\mathcal{N}_K(u) = \operatorname{TopK}_{k}\; \mathrm{sim}(u,z_k).
\end{equation}
The retrieved contexts can then be used to augment the current input before multimodal reasoning. This retrieval mechanism provides a non-parametric way to condition the model on relevant past observations or semantically related examples.

\subsection{Policy Optimization}

Given the latent state $h$, a policy head samples an action from distribution over candidate answers $a\sim\pi_{\phi}(\cdot \mid h),$
where $\phi$ denotes the policy parameters, and receives a corresponding reward $r(h,a)$. For multiple-choice VQA, the selected action corresponds to one of the answer candidates in $\mathcal{A}$.

The objective is to maximize the expected reward $J(\pi_{\phi}) = \mathbb{E}_{\pi_{\phi}}[r(h,a)].$
Using the policy gradient theorem, the policy parameters can be optimized via
\begin{equation}
\nabla_{\phi} J(\pi_{\phi})
=
\mathbb{E}_{\pi_{\phi}}
\left[
\nabla_{\phi}\log \pi_{\phi}(a \mid h)\,
A^{\pi_{\phi}}(h,a)
\right],
\end{equation}
where $A^{\pi_{\phi}}(h,a)$ denotes the advantage function.} 

\section{Methodology}

\subsection{Belief-aware contextualization}

We formulate our problem under partial observability, where the   VLM $\pi_{\theta}$ must reason the latent human intent  from multi-modal observations. The interaction evolves over time, where at each time step $t$, the agent observes
$o(t) = \{v(t), \ell(t)\}$
where $v(t)$ denotes visual observations and $\ell(t)$ denotes contextual language inferred from the scene. 
We assume that the intent $q \in \mathcal{Q}$ for the human is latent and evolves on a \emph{slow time-scale} and remains constant during human-environemnt sensorimotor interaction. $\mathcal{Q}$ represents set of all possible intents for agent.
\begin{equation}
q(\tau + 1) \approx q(\tau), \quad \forall t \in [\tau, \tau + \Delta T]
\end{equation}

Further, unlike prior approaches that learn belief dynamics through an explicit parametric update rule, our formulation estimates belief by retrieving relevant past experiences from memory. Inspired by human cognition and Theory of Mind~\cite{pmlr-v80-rabinowitz18a}, we assume that decision-making can be guided by retrieving similar situations rather than modeling belief transitions with a predefined recursion. Motivated by this, we construct a belief representation by freezing the multi-modal fusion layers of the VLM and use them to encode each observed interaction into a latent representation.  We build a memory of these past multi-modal latent representations $\mathcal{M} = \{(z_k, a_k, r_k)\}_{k=1}^{N}$.
Concretely, the vector-based memory of vision–language embeddings can be used to approximate the belief by retrieving the top-
K
most relevant contexts based on similarity.

\begin{figure}
    \centering  \includegraphics[width=0.9\linewidth]{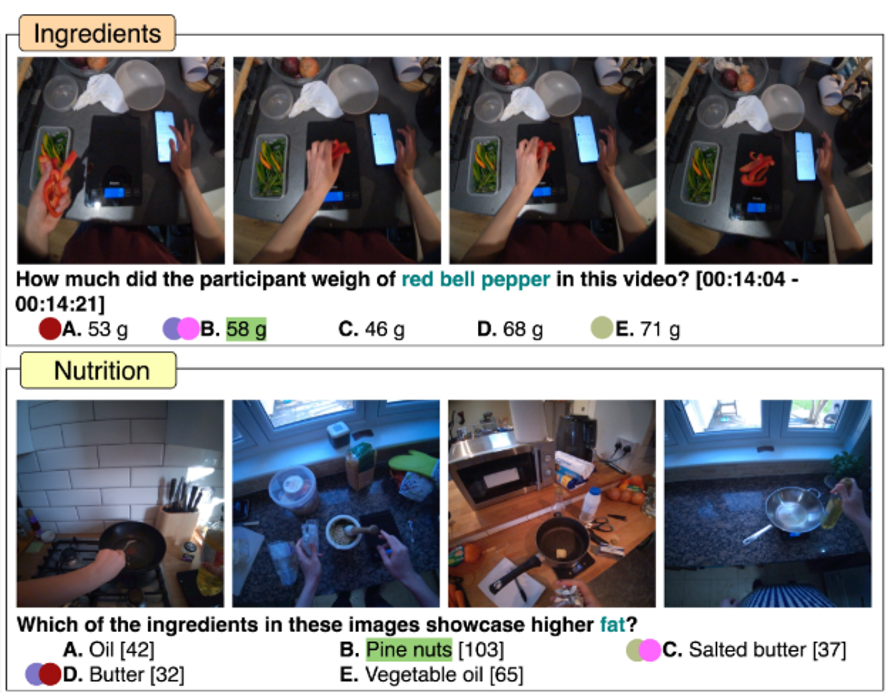}
    \caption{VQA samples from HD-EPIC with egocentric video frames, questions, and answer choices.}
    \label{fig:HD_Epic}
\end{figure}

To retrieve the top-$K$ contextually similar embeddings, we encode the visual observation $v(t)$ and language input $\ell(t)$ using pretrained vision and language encoders, respectively, and form the joint latent representation
$
z(t)=\!\big(\phi_{\mathrm{vis}}(v(t)),\,\phi_{\mathrm{lang}}(\ell(t))\big)
$.  We can now retrieve the top-$K$ most similar entries 
\begin{equation}
\mathcal{N}_K(z(t)) = \text{TopK}_k \; \text{sim}(z(t), z_k),
\end{equation}
where $\text{sim}(\cdot,\cdot)$ denotes a similarity function (e.g., cosine similarity), $z_{k} \in \mathcal{M} $ are the embeddings from the memory bank. The belief is computed by aggregating the top-$K$ retrieved memory entries using similarity-based importance weights:
\begin{align}
b(t) &= \sum_{k \in \mathcal{N}_K(z(t))} w_k \, c_k, \\ \nonumber
w_k &= \frac{\exp(\mathrm{sim}(z(t), z_k))}{\sum_{j \in \mathcal{N}_K(z(t))} \exp(\mathrm{sim}(z(t), z_j))},
\end{align}
where $c_k$ denotes the contextual content associated with memory embedding $z_k$. In practice, the top-$K$ retrieved contexts are assigned importance scores based on their similarity to the current embedding $z(t)$, and these scores are used to pool the retrieved contexts into a single belief vector $b(t)$. This retrieval process yields an implicit belief representation without requiring direct supervision. The prior belief $b(t)$ retrieved from the vector database, together with the current latent embedding $z(t)$, is then passed to the VLM backbone to obtain the belief-augmented, task-contextualized representation $h(t)$, which serves as the latent state for downstream action selection.

\begin{equation}
h(t) = \pi_{\theta}(z(t), b(t))
\end{equation}

In practice, $h(t)$ is extracted as the final layer of the VLM backbone. 
 Training can be performed either via full fine-tuning of $\theta$ or through parameter-efficient fine-tuning (PEFT) methods. Typically, action selection is modeled as a direct mapping from observations and prior belief to actions:
\begin{equation}
a(t) \sim \pi_\theta(\cdot \mid h(t))
\end{equation}





\subsection{RL for human reasoning}

While this formulation enables belief-aware contextualization, it does not explicitly optimize decision-making with respect to task objectives, which can lead to suboptimal action selection. In particular, the retrieval-based belief representation captures relevant past context but lacks a mechanism to refine actions based on long-horizon outcomes. To address this, we further optimize the policy using reinforcement learning, where task-specific rewards guide the model to improve action selection conditioned on both current observations and inferred belief. This enables the model to align its predictions with downstream objectives and enhances its ability to perform consistent, goal-directed reasoning.

\begin{figure*}
    \centering  \includegraphics[width=1.0\linewidth]{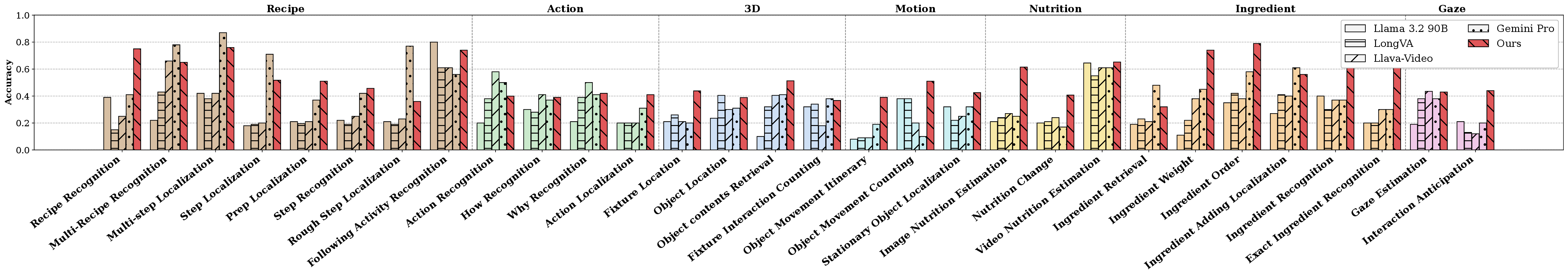}
    \caption{\textbf{VQA performance across diverse question prototypes and categories (recipe, action, 3D, motion, nutrition, ingredient, and gaze).}}
    \label{fig:VQA}
\end{figure*}

We define a policy $\pi_{\phi} : \mathbb{R}^{d} \rightarrow \mathcal{A}$ that takes the belief-aware representation $h(t)$ as input and outputs a distribution over actions:
\begin{equation}
a \sim \pi_{\phi}(\cdot \mid h(t)),
\end{equation}
where $\phi$ denotes the parameters of a lightweight policy network.
Based on the correctness of the predicted action $a$,
we define a binary reward
\begin{equation}
r(h,a)=
\begin{cases}
1, & \text{if } a=y,\\
0, & \text{otherwise}.
\end{cases}
\end{equation}

We fine-tune the policy using Proximal Policy Optimization (PPO). The PPO objective is given by
\begin{equation}
\mathcal{L}_{\text{PPO}} = \mathbb{E}_t \left[
\min \left(
r_t(\phi) \hat{A}_t,\;
\text{clip}(r_t(\phi), 1-\epsilon, 1+\epsilon)\hat{A}_t
\right)
\right],
\end{equation} \label{eq:ppo_loss}
where
\begin{equation}
r_t(\phi) = \frac{\pi_{\phi}(a_t \mid h(t))}{\pi_{\phi_{\text{old}}}(a_t \mid h(t))}
\end{equation}
is the probability ratio between the updated and previous policies, $\hat{A}_t$ denotes the estimated advantage function, and $\epsilon$ is a clipping parameter that stabilizes training.
This objective encourages the policy to improve action selection while preventing large deviations from the previous policy, ensuring stable and efficient learning.

\medskip

\section{Experiments}

We evaluate our approach on video question answering (VQA) settings, where models are required to reason over egocentric video context and select the correct answer from multiple choices. We train our belief-aware VLM on HD-EPIC and compare its performance against strong vision-language baselines. Our goal is to assess whether incorporating belief improves reasoning and action understanding in complex, multimodal scenarios.

\subsection{HD-EPIC Dataset}

The HD-EPIC dataset is an  egocentric video dataset that provides rich multimodal annotations for fine-grained human activity understanding. As shown in Fig.~\ref{fig:HD_Epic}, the dataset consists of:

\begin{itemize}
    \item \textit{Egocentric video:} First-person high-resolution video capturing human-object interactions.
    \item \textit{Action annotations:} Fine-grained verb-noun pairs representing human actions.
    \item \textit{Eye-gaze signals:} Attention cues indicating regions of interest, providing implicit supervision for intent.
    \item \textit{Temporal context:} Long-horizon sequences enabling modeling of action progression and task structure.
\end{itemize}

These modalities provide complementary signals for understanding both observable actions and underlying intent. The reasoning model requires long horizon contextual understanding of the scene to predict human intent, which current baselines struggle to acheive.

\subsection{Baselines}

We compare our method against the following state-of-the-art vision-language models:

\begin{itemize}
    \item \textit{Llama 3.2 90B:} A large-scale multimodal foundation model with strong reasoning capabilities.
    \item \textit{LongVA:} Longest context open-source model designed to capture temporal dependencies in extended video sequences.
    \item \textit{LLaVA-Video:} A video-language model that extends LLaVA to handle temporal visual inputs.
    \item \textit{Gemini-Pro:} Closed source, longest context of any model, and state-of-the-art on long-video.
\end{itemize}

These baselines represent diverse approaches to video understanding, ranging from general-purpose large models to specialized long-horizon reasoning architectures.

\begin{table}[t]
\centering
\scriptsize
\setlength{\tabcolsep}{2.5pt}   
\renewcommand{\arraystretch}{1.05}

\resizebox{\columnwidth}{!}{ 
\begin{tabular}{lcccccccc}
\toprule
\textbf{Model} & 
\cellcolor{red!20}\textbf{Recipe} & 
\cellcolor{orange!25}\textbf{Ingredient} & 
\cellcolor{yellow!30}\textbf{Nutrition} & 
\cellcolor{green!25}\textbf{Action} & 
\cellcolor{blue!20}\textbf{3D} & 
\cellcolor{cyan!20}\textbf{Motion} & 
\cellcolor{magenta!20}\textbf{Gaze} & 
\textbf{Avg.} \\
\midrule

\multicolumn{9}{l}{\textbf{Blind - Language Only}} \\
Llama 3.2 & 33.5 & 25.0 & 36.7 & 23.3 & 22.3 & 25.5 & 19.5 & 26.5 \\
Gemini Pro & 38.0 & 26.8 & 30.0 & 22.1 & 21.5 & 27.7 & 20.5 & 26.7 \\

\midrule
\multicolumn{9}{l}{\textbf{Video-Language}} \\
VideoLlama 2 & 30.8 & 25.7 & 32.7 & 27.2 & 25.7 & 28.5 & 21.2 & 27.4 \\
LongVA & 29.6 & 30.8 & 33.7 & 30.7 & 32.9 & 22.7 & 24.5 & 29.3 \\
LLaVA-Video & 36.3 & 33.5 & 38.7 & \textbf{43.0} & 27.3 & 18.9 & 29.3 & 32.4 \\
Gemini Pro & \textbf{60.5} & 46.2 & 34.7 & 39.6 & 32.5 & 20.8 & 28.7 & 37.6 \\

\midrule
\multicolumn{9}{l}{\textbf{Ours}} \\
{VLM + RL}  & 49.5 & 51.7 & 53.2 & 40.5 & 40.2 & 38.6 & \textbf{43.5} & \textbf{44.6} \\

{Belief + VLM + RL}  & 54.9 & \textbf{59.2} & \textbf{56.8} & \textbf{43.0} & \textbf{42.1} & \textbf{42.3} & \textbf{43.5} & \textbf{48.8} \\

\midrule
\textit{Sample Human Baseline} & 96.7 & 96.7 & 85.0 & 92.5 & 93.8 & 92.7 & 75.0 & 90.3 \\
\bottomrule
\end{tabular}
}

\caption{\textbf{VQA Results per Category (\% Acc.).} Our benchmark remains comparable to state-of-the-art  VLM models.}
\label{tab:vqa_results}
\end{table}

\subsection{Training}

We train the belief-aware VLM by adopting InternVL-2B \cite{chen2024internvl} as the backbone  architecture.  We compute the cross-entropy (CE) loss for VLM fine-tuning 
\begin{equation}
\mathcal{L}_{\text{CE}} = - \sum_t \log p(y_t \mid h(t)),
\end{equation}
and the policy gradient loss Eq.~\ref{eq:ppo_loss} for the policy head. The overall network is trained against a cumulative loss  
\[
\mathcal{L} = w_1 \,\mathcal{L}_{CE}+w_2\,\mathcal{L}_{PPO}.
\]
We train across eight NVIDIA H200 GPUs using distributed data-parallel technique.  From the video clip, we uniformly sample 8 frames for processing with a batch size of 4 and gradient accumulation of 64. 








\section{Results}

Table~\ref{tab:vqa_results} provides a quantitative comparison demonstrating that explicitly incorporating human belief into VLMs significantly improves their reasoning capabilities, resulting in more consistent and human-aligned decisions. We first establish a baseline by augmenting a
VLM with a reinforcement learning policy (VLM + RL), where the model reasons over multimodal
inputs (video and text) and is optimized using reward signals derived from action correctness. This
baseline captures task-relevant perception and decision-making but lacks an explicit mechanism to
model prior belief or agent intent. To address this, we extend the architecture by incorporating a
retrieval-based belief module (VLM + RL + Belief), which queries a vector memory of multi-modal
embeddings to retrieve top-K similar past contexts and forms belief context via similarity-weighted
aggregation. 
These belief tokens are integrated into the VLM, enabling the model to condition its
reasoning on prior experience and inferred intent.

We evaluate both models on the HD-EPIC \cite{11094504} dataset, which provides rich multimodal annotations including egocentric video, eye gaze, transcriptions, and verb–noun action labels. We construct a VQA-style evaluation setting, where the model is given a video segment and a query with multiple candidate answers, and performance is measured based on correctness accuracy across diverse categories such as \textit{Recipe, Ingredient, Nutrition, Action, 3D Perception, Motion, and Gaze}. Our results show that incorporating belief leads to consistent improvements across all categories, with an overall gain of +4\% compared to the VLM + RL baseline. Notably, the largest improvements are observed in multi-step and context-dependent categories such as \textit{3D Perception}, \textit{Ingredient}, and \textit{Nutrition}, where reasoning over prior context and intent is critical. The proposed VLM + RL + Belief model achieves the best overall accuracy of \textbf{48.8\%}, demonstrating its ability to capture both perceptual and cognitive aspects of human behavior.

We further compare our method against strong baselines, including LLaVA-Video, LongVA, and Gemini Pro. Compared to LLaVA-Video (32.4\%) and LongVA (29.3\%), our model achieves a significant improvement of over +10\% absolute gain, demonstrating the effectiveness of combining belief-aware reasoning with reinforcement learning. Notably, our approach also outperforms large-scale models such as Gemini Pro (37.6\%), while maintaining a significantly smaller model size ($\sim$2.37B parameters), with the policy network $\pi_{\phi}$ comprising only $\sim$8.4M parameters. his highlights the effectiveness of belief-aware modeling in enabling efficient
and scalable reasoning, even with substantially fewer parameters compared to large-scale foundation
models.

\medskip

We further analyze performance across different question types, including \textit{Recipe, Ingredient, Nutrition, Action, 3D, Motion}, and \textit{Gaze}.

\begin{itemize}
    \item \textit{Ingredient and Nutrition:}  
    Our model performs best on Ingredient (59.2\%) and Nutrition (56.8\%), where belief-aware context helps distinguish similar visual cues.

    \item \textit{Action and Motion Understanding:}  
    Strong gains on Action (43.0\%) and Motion (42.3\%) suggest improved modeling of temporal dynamics and human intent.

    \item \textit{3D and Spatial Reasoning:}  
    Our method achieves 42.1\% on 3D tasks, indicating better handling of spatial relationships and scene consistency.

    \item \textit{Gaze Estimation:}  
    The largest gain appears in Gaze (43.5\%), showing the model’s strength in inferring attention and intent.
\end{itemize}

\medskip







\addtolength{\textheight}{-2cm}   





\section{Conclusion}

We presented a belief-aware vision--language model for egocentric video reasoning that explicitly incorporates latent belief to capture intent and contextual dependencies. By leveraging a retrieval-based memory of past multimodal experiences and refining the policy with reinforcement learning, our approach enables more consistent and goal-directed decision-making compared to standard VLMs. Experiments on the HD-EPIC dataset demonstrate significant improvements over strong baselines across diverse VQA tasks, particularly in scenarios requiring temporal reasoning and intent inference. These results highlight the importance of modeling belief for human-like reasoning and suggest a promising direction for integrating cognitive principles into large multimodal models.

\printbibliography

\end{document}